\documentclass[10pt,twocolumn,letterpaper]{article}

\usepackage[pagenumbers]{cvpr} % To force page numbers, e.g. for an arXiv version

\usepackage{graphicx}
\usepackage{amsmath}
\usepackage{amssymb}
\usepackage{enumitem}
\usepackage{algorithm}
\usepackage{algorithmicx}
\usepackage{booktabs}
\usepackage{xcolor, colortbl}
\usepackage{adjustbox}
\usepackage{tabularx}
\usepackage{multirow}
\usepackage{array}
\usepackage{marvosym}

\usepackage[pagebackref,breaklinks,colorlinks]{hyperref}

\usepackage[capitalize]{cleveref}
\crefname{section}{Sec.}{Secs.}
\Crefname{section}{Section}{Sections}
\Crefname{table}{Table}{Tables}
\crefname{table}{Tab.}{Tabs.}

\begin{document}

\title{From Node Interaction to Hop Interaction: New Effective and Scalable Graph Learning Paradigm}

\author{Jie Chen$^{1}$\qquad Zilong Li$^{1}$\qquad Yin Zhu$^{1}$\qquad Junping Zhang$^{1}$\qquad Jian Pu$^{2*}$ \\
$^{1}$ Shanghai Key Lab of Intelligent Information Processing, School of Computer Science,\\
Fudan University, Shanghai 200433, China\\
$^{2}$ Institute of Science and Technology for Brain-Inspired Intelligence,\\  Fudan University, Shanghai 200433, China\\
{\tt\small \{chenj19, yinzhu20, jpzhang, jianpu\}@fudan.edu.cn}, \tt\small zilongli21@m.fudan.edu.cn
}

 \maketitle

%%%%%%%%% ABSTRACT
\begin{abstract}
Existing Graph Neural Networks (GNNs) follow the message-passing mechanism that conducts information interaction among nodes iteratively. While considerable progress has been made, such node interaction paradigms still have the following limitation. 
First, the scalability limitation precludes the broad application of GNNs in large-scale industrial settings since the node interaction among rapidly expanding neighbors incurs high computation and memory costs. 
Second, the over-smoothing problem restricts the discrimination ability of nodes, i.e., node representations of different classes will converge to indistinguishable after repeated node interactions. In this work, we propose a novel hop interaction paradigm to address these limitations simultaneously. 
The core idea is to convert the interaction target among nodes to pre-processed multi-hop features inside each node.
We design a simple yet effective HopGNN framework that can easily utilize existing GNNs to achieve hop interaction. Furthermore, we propose a multi-task learning strategy with a self-supervised learning objective to enhance HopGNN. We conduct extensive experiments on 12 benchmark datasets in a wide range of domains, scales, and smoothness of graphs. Experimental results show that our methods achieve superior performance while maintaining high scalability and efficiency. The code is at~\url{https://github.com/JC-202/HopGNN}.
\end{abstract}

\section{Introduction}
Graph Neural Networks (GNNs) have recently become very popular 
and have demonstrated great results in a wide range of graph applications, including social networks~\cite{Sun_2020_CVPR}, point cloud analysis~\cite{shi2020point} and recommendation systems~\cite{he2020lightgcn}.
The core success of GNNs lies in the message-passing mechanism that iteratively conducts information interaction among nodes~\cite{gilmer2017neural,gasteigerdirectional,chen2022memory}. 
Each node in a graph convolution layer first aggregates information from local neighbors and combines them with non-linear transformation to update the self-representation~\cite{GCN, GAT, Hamilton2017InductiveRL}.
After stacking $K$ layers, nodes can capture long-range $K$-hop neighbor information and obtain representative representations for downstream tasks~\cite{xu2018representation,li2021deepgcns}.
However, despite the success of such popular node interaction paradigms, the number of neighbors for each node would grow exponentially with layers~\cite{alon2020bottleneck,topping2022understanding}, resulting in the well-known \textit{scalability} and \textit{over-smoothing} limitation of GNNs.

\begin{figure}[!t]
	\centering
	\includegraphics[width=0.9\linewidth]{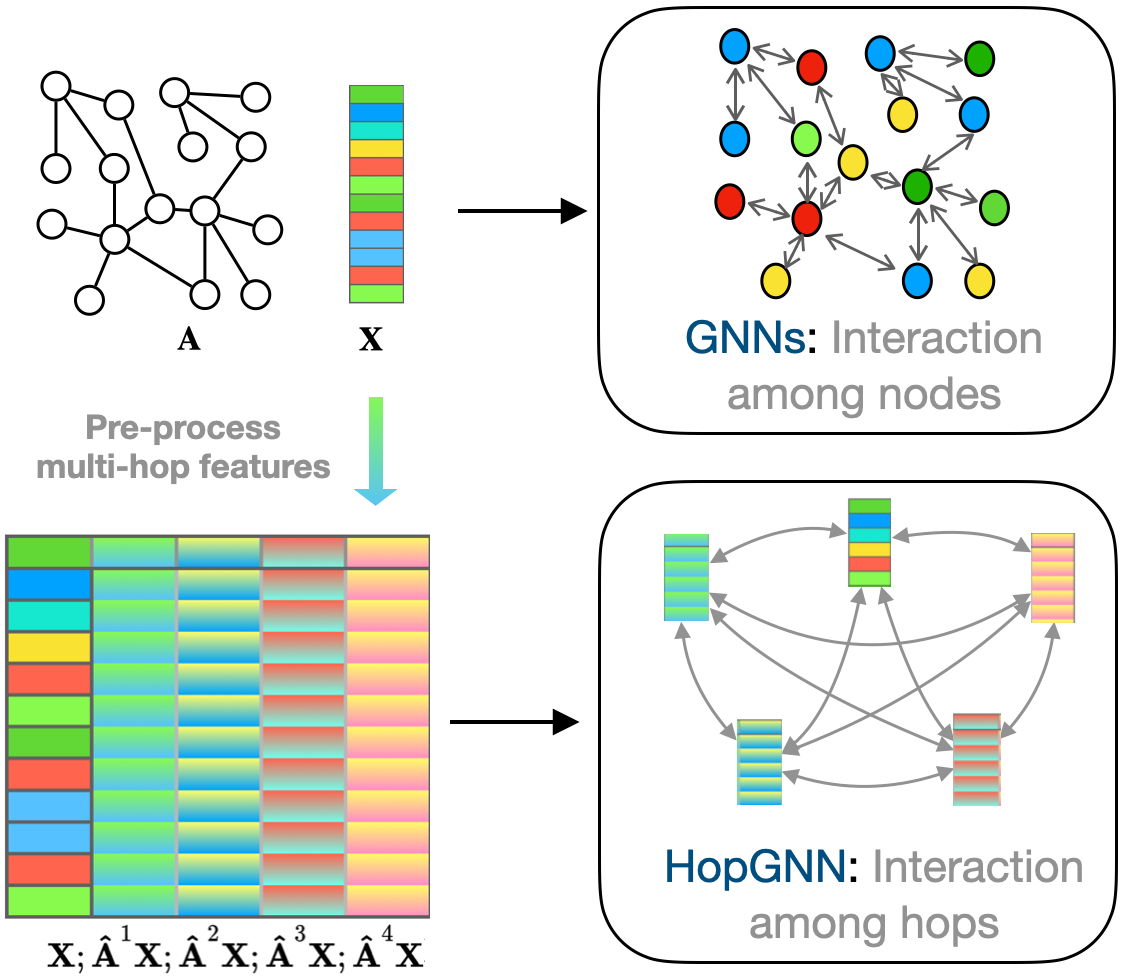}
	\caption{Comparison of node interaction and hop interaction. 
The hop interaction first pre-computes multi-hop features and then conducts non-linear interaction among different hops via GNNs, which enjoy high efficiency and effectiveness.
    }
	\label{fig:intro}
\end{figure}

%\textbf{Limitation 1:} 
The \textit{scalability} limitation precludes the broad application of GNNs in large-scale industrial settings since the node interaction among rapidly expanding neighbors incurs high computation and memory costs~\cite{zhang2021graph,duan2022comprehensive}.
Although we can reduce the size of neighbors by sampling techniques~\cite{Hamilton2017InductiveRL, chen2018fastgcn}, it still executes node interaction iteratively during training, and the performance is highly sensitive to the sampling quality~\cite{duan2022comprehensive}.
Recently, scalable GNNs that focus on simplifying or decoupling node interactions have emerged~\cite{wu2019simplifying, frasca2020sign,zhu2020simple}. 
Such decoupled GNNs first pre-compute the linear aggregation of K-hop neighbors to generate node features and then utilize the MLP to each node without considering the graph structure during training and inference. However, despite high efficiency and scalability, such methods lead to suboptimal results due to the lack of nonlinear interactions among nodes.

%\textbf{Limitation 2:} 
Another limitation is \textit{over-smoothing}, which restricts the discriminative ability of nodes, \ie, node representations will converge to indistinguishable after repeated node interactions~\cite{li2018deeper,chen2020measuring}.
On one hand, it causes performance degeneration when increasing layers of GNNs~\cite{GCN,liu2020towards}. On the other hand, in some heterophily graphs where connected nodes are usually from different classes, shallow GNNs are also surprisingly inferior to pure Multi-ayer Perceptrons (MLPs)~\cite{pei2019geom,zhu2020beyond}. The reason is the interaction among massive local inter-class neighbors would blur class boundaries of nodes~\cite{zhu2020beyond,hou2020measuring,chen2022graph}. 
Recently, to carefully consider the neighbor influence and maintain the node discrimination, emerging advanced node interaction GNNs, such as deep GNNs with residual connections~\cite{chen2020simple,li2021deepgcns} and heterophilic-graph-oriented GNNs with adaptive aggregation~\cite{bo2021beyond,suresh2021breaking,yan2021two,luan2021heterophily}, have achieved promising results.
However, these advanced node interactions suffer high computational costs and fail to handle large-scale datasets.

%\textbf{Motivation:} 
These two limitations have typically been studied separately, 
 as addressing one often necessitates compromising the other.
However, \textit{can we bridge the two worlds, enjoying the low-latency, node-interaction-free of decoupled GNNs and the high discrimination ability of advanced node interaction GNNs simultaneously?} We argue that it is possible to transform the node interaction into a new hop interaction paradigm without losing performance, but drastically reducing the computational cost. As shown in Figure~\ref{fig:intro}, the core idea of hop interaction is to decouple the whole node interaction into two parts, the non-parameter hop feature pre-processing and non-linear interaction among hops.
Inspired by the recommendation system, the non-linear interaction among different semantic features can enhance discrimination~\cite{Guo2017DeepFMAF}, \eg, model the co-occurrence of career, sex and age of a user to identify its interest.
By treating the precomputed $L$ hop neighbors as $L$ semantic features within each node, we can consider node classification as a feature interaction problem, \ie, model the non-linear hop interaction to obtain discriminative node representations.

To this end, we design a simple yet effective HopGNN framework to address the above limitation simultaneously. It first pre-computes the multi-hop representation according to the graph structure. Then, without loss of generality, we can utilize GNNs over a multi-hop feature graph inside each node to achieve hop interaction flexibly and explicitly.
Specifically, we implement an attention-based interaction layer and average pooling for the HopGNN to fuse multi-hop features and generate the final prediction.
Furthermore, to show the generality and flexibility of our framework, we provide a multi-task learning strategy that combines the self-supervised objective to enhance performance. 

Our contributions are summarized as follows: 

%\begin{itemize}
1. \textit{New perspective:} 
We propose a new graph learning paradigm going from node to hop interaction. It conducts non-linear interactions among pre-processed multi-hop neighbor features inside each node.

2. \textit{General and flexible framework:} We design a simple yet effective HopGNN framework for hop interaction. Besides, the HopGNN is general and flexible to combine the self-supervised objective to easily enhance performance.
    
3. \textit{State-of-the-art performance:}  Experimental results show HopGNN achieves state-of-the-art performance on 12 graph datasets of diverse domains, sizes and smoothness while maintaining high scalability and efficiency.
%\end{itemize}

%\input{sections/Preliminary}
\section{Background and Related Works}
\subsection{Notation and Node Classification Problem}
%\noindent \textbf{Notation and Node Classification. }
Consider a graph ${\mathcal{G} = (\mathcal{V}, \mathcal{E})}$, with $N$ nodes and $E$ edges. Let ${\mathbf{A}\in \mathbb{R}^{N\times N}}$ be the adjacency matrix, with $\mathbf{A}_{i,j} = 1$ if edge$(i, j) \in \mathcal{E}$, and 0 otherwise. 
$\mathbf{X} \in \mathbb{R}^{N\times d}$ and $\mathbf{Y} \in \mathbb{R}^{N\times c}$ represent the features and labels of nodes, repsectively.
Given a set of labeled nodes ${\mathcal{V_L}}$, the task of node classification is to predict the labels of the unlabeled nodes by exploiting the graph structure $\mathbf{A}$ and features $\mathbf{X}$.

Besides, the notion of homophily and heterophily corresponds to the smoothness of the signal $\mathbf{Y}$ on the graph $\mathcal{G}$. The edge homophily ratio of the graph is defined as $\textstyle{\mathcal{H}_\text{edge}=\frac{\left|\left\{(u, v):(u, v) \in \mathcal{E} \wedge y_u=y_v\right\}\right|}{|\mathcal{E}|}}$, while $\mathcal{H}_\text{edge}$ tends to 1 means high homophily and low heterophily, and vice versa.

\subsection{Graph Neural Networks}

\noindent \textbf{Standard node interaction GNNs.}
Each layer of most node interaction GNNs follows the message-passing mechanism~\cite{gilmer2017neural} that is composed of two steps: (1) aggregate information from neighbors: $\textstyle{\mathbf{m}_i^l = \operatorname{AGGREGATE}(\mathbf{h_j}^l, v_j \in \mathcal{N}_i)}$; (2) update representation: $\textstyle{\mathbf{h_i}^{l+1} = \operatorname{UPDATE}(\mathbf{h_i}^{l}, \mathbf{m}_i^l)}$.
% Without loss of generality, most wildely used full-batch GNNs implement these two steps in the form as following:
To capture long-range information, standard node interaction GNNs alternately stack graph convolutions, linear layers and non-linear activation functions to generate representative representations of nodes~\cite{tnnls-compre}. Without loss of generality, the widely used GNNs follow the following form:
\begin{align}
\mathbf{H}^{L}=\mathbf{{\hat{A}}} \ \sigma\left(\cdots \sigma\left(\mathbf{\hat{A}} \mathbf{X}\mathbf{W}\right) \cdots\right)  \mathbf{W}^{L-1}
\end{align} 
where $\mathbf{\hat{A}}$ is a normalized weighted matrix for feature propagation, $\mathbf{W}^{l}$ is a learnable weight matrix and $\sigma$ is a non-linear activation function $\operatorname{Relu}$. For example, the GCN~\cite{GCN} utilizes the symmetric normalized adjacent matrix as $\mathbf{\hat{A}}$, GraphSAGE~\cite{Hamilton2017InductiveRL} utilizes the random walk normalized version, and the GAT~\cite{GAT} applies the attention mechanism~\cite{vaswani2017attention} to obtain a learnable weighted $\mathbf{\hat{A}}$.

\noindent \textbf{Advanced node interaction GNNs.} 
To avoid over-smoothing and heterophily problems that arise from simply stacking graph convolution layers~\cite{li2018deeper,pei2019geom}, most advanced node interaction GNNs adopt residual connections~\cite{li2019deepgcns,xu2018representation} or adaptive aggregation strategies~\cite{pei2019geom,zhu2020beyond,li2022finding,chen2022exploiting} to extend standard GNNs. For instance, GCNII~\cite{chen2020simple} is the SOTA deep GNN that combines GCN with initial connection and identity mapping. 
Geom-GCN~\cite{pei2019geom} and WRGAT~\cite{suresh2021breaking} transform the original graph by discovering the non-local semantic similarity neighbor.
H2GCN~\cite{zhu2020beyond} utilizes ego and neighbor separation and higher-order combination to improve the performance of GNNs under heterophily.
GGCN~\cite{yan2021two} adopt signed messages from nodes' local neighbors and a degree correction mechanism for node-wise rescaling.
FAGCN~\cite{bo2021beyond} and ACM-GCN~\cite{luan2021heterophily} apply low-pass and high-pass filters in each graph convolution layer.
Despite the high expressiveness of these advanced GNNs, they suffer from high computational costs and limited scalability to large-scale graphs.

\noindent \textbf{Sampling-based GNNs. }
Sampling-based GNNs reduce memory consumption by sampling and minibatch training to approximate the full-batch GNNs. There are three categories of widely-used sampling strategies: 1) \textit{Node-wise sampling}, GraphSAGE~\cite{Hamilton2017InductiveRL} randomly samples a fixed-size set of neighbors for each node in every layer, and VR-GCN ~\cite{chen2018stochastic} further analyzes the variance reduction for node sampling. 2) \textit{Layer-wise sampling}, Fast-GCN~\cite{chen2018fastgcn} samples a fixed number of nodes at each layer, and ASGCN~\cite{huang2018adaptive} proposes adaptive layer-wise sampling with better variance control. 3) \textit{Graph-wise sampling}, ClusterGCN~\cite{chiang2019cluster} first partitions the entire graph into clusters and then samples the nodes in the clusters, and GraphSAINT~\cite{zeng2019graphsaint} directly samples a subgraph for mini-batch training. However, these sampling strategies still conduct node interactions that face high communication costs during training, and the model performance is highly sensitive to sampling quality~\cite{duan2022comprehensive}.

\noindent \textbf{Decoupled GNNs. }
Unlike standard GNNs, which sequentially stack non-linear graph convolution layers, decoupled GNNs simplify the graph convolution by decoupling the model into two steps: hop feature propagation and an MLP classifier for prediction. Depending on the propagation order, there are two typical ways to decouple these two operations: (1) \textit{Pre-processing} that first pre-computes the feature propagation, \eg, $\mathbf{X}=\sum_{i=0}^L \theta_i\mathbf{\hat{A}}^i\mathbf{X}$, and then applies the MLP classifier for each node representation $\mathbf{X}^L$ individually. 
SGC~\cite{wu2019simplifying} simplifies GNNs into a linear model $\mathbf{Y}=\mathbf{\hat{A}}^L\mathbf{X}\mathbf{W}$, which achieves faster computation. However, it only considers the $L$ hop features, and the linear layer limits its expressiveness. S2GC~\cite{zhu2020simple} extends the SGC by using the simple spectral graph convolution to average the propagated features in multi-hop. It also proposes utilizing MLP as a classifier. Different from S2GC, SIGN~\cite{frasca2020sign} uses the concatenation of multi-hop features, with an individual W for each hop transformation, to achieve better performance. 
(2) \textit{Post-processing} that propagates multi-hop features after an MLP predictor: $\mathbf{Y}=\textstyle{\sum_{i=0}^L \theta_i}\mathbf{\hat{A}}^i\operatorname{MLP}(\mathbf{X})$. To this end, APPNP~\cite{klicpera2018predict} utilizes the approximate personal PageRank filter as a feature propagator, while GPRGNN~\cite{chien2020adaptive} further utilizes a learnable generalized PageRank for the weights of theta to enhance performance. However, such post-processing still needs to propagate high-cost features during training, limiting its scalability compared to pre-processing. 

Although decoupled GNNs efficiently propagate features, they usually suffer suboptimal results in heterophilic graphs due to the lack of non-linear node interactions. 
Our proposed approach builds on the efficiency of pre-computed hop features, as in decoupled GNNs.
However, we introduce a novel paradigm that explicitly considers the non-linear interactions among hops, which is more expressive and discriminative without sophisticated node interactions.

\section{Methodology}
\label{sec:method}

\begin{figure*}[t]
	\centering
	\includegraphics[width=0.95\linewidth]{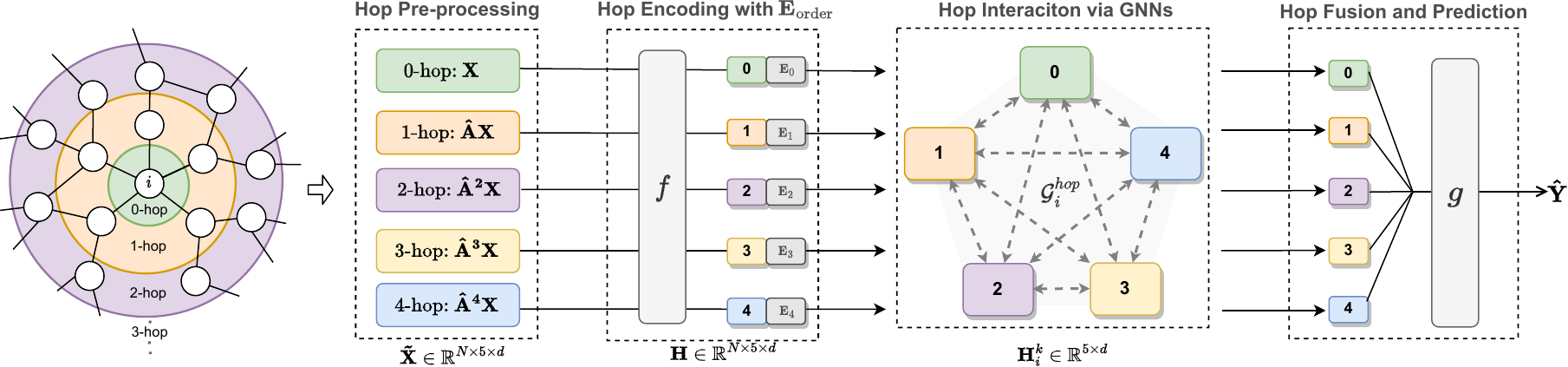}
	\caption{Overview of the proposed HopGNN framework with four steps, illustration with four hops. The core idea of HopGNN is to convert the interaction target of standard GNNs from nodes to pre-processed multi-hop features inside each node, which can enhance the nodes' discriminative power without node interaction.}
	\label{fig:hopgnn}
\end{figure*}

\subsection{HopGNN}
As illustrated in Figure~\ref{fig:hopgnn}, our proposed HopGNN framework is composed of four steps, namely, hop preprocessing, hop encoding, hop interaction, hop fusion and prediction. We introduce them one by one in detail. 

\noindent \textbf{Hop Pre-processing.}
Extracting information from long-range neighbors is crucial for nodes to achieve representative representation~\cite{xu2018representation,abu2019mixhop}.
Unlike the node interaction GNNs, which stack $L$ layers to reach $L$-hop neighbors and suffer high computational costs with limited scalability~\cite{zhang2021graph}, our proposed framework is built upon a hop-information pre-processing step. It pre-computes multi-hop node information from the node's original feature vectors and the $l$-hop feature can be simplified as:
\begin{align}
\tilde {\mathbf{X}} = [\mathbf{X}_p;\mathbf{X}_p^1...;\mathbf{X}_p^L], \quad \mathbf{X}_p^l =\mathbf{
\hat{A}}^l \mathbf{X},
\end{align}
where $\mathbf{\hat{A}}$ is the normalized adjacent matrix, and $\tilde {\mathbf{X}}$ contains the multi-hop neighbor information. This pre-processing does not require any learnable parameters and only needs to be computed once in a CPU or distributed system for large-scale graphs. As a result, it can make models naturally support mini-batch training and easily scale to large datasets since it eliminates the computational complexity of node aggregation during training and inference. 

\noindent \textbf{Hop Encoding.}
To obtain sufficient expressive power, it is necessary to have at least one learnable linear transformation to transform the input hop features into higher-level hop embedding.
For parameter efficiency, we apply a shared parametered linear layer $\operatorname{f}$ to encode all hops.
Moreover, to incorporate the semantic order information of the hops, we add a 1D learnable hop-order encoding vector $\mathbf{E}_{\text {order}} \in \mathbb{R}^{1\times L \times d}$ to each node hop embedding.
\begin{align}
    \mathbf{H}=\left[\operatorname{f}(\mathbf{X}_p) ; \operatorname{f}(\mathbf{X}_p^1); \cdots ; \operatorname{f}(\mathbf{X}_p^L)\right]+\mathbf{E}_{\text {order}}.
\end{align}
The hop embedding $\mathbf{H} \in \mathbb{R}^{N \times L \times d}$ contains multi-hop neighbor information. The $\mathbf{E}_{\text {order}}$ can help the following order-insensitive hop interaction layer to capture hop-order information, which is important for the heterophily datasets, as further discussed in Section~\ref{sec:ablation}.

\noindent \textbf{Hop Interaction.}
Inspired by~\cite{Guo2017DeepFMAF}, we argue that the underlying co-occurrence among different hop features inside each node contains the clue for its discrimination. Therefore, our goal is to model the non-linear interaction among hops to enhance discrimination without node interaction. 
Since message-passing is widely used in GNNs for non-linear interactions among nodes~\cite{gilmer2017neural}, to achieve hop interaction, we take advantage of well-designed GNNs by shifting the interaction target from nodes to multi-hop features inside each node.
Specifically, for node $v_i$, we construct a fully-connected hop feature graph $\mathcal{G}_i^\text{hop}$, where each node $v'_l$ in $\mathcal{G}_i^\text{hop}$ corresponds to an $l$-th hop feature of the original node $v_i$. 
Then, by leveraging any standard GNNs to model node interactions on the feature graph $\mathcal{G}_i^\text{hop}$, we can effectively capture multi-hop feature interactions of node $v_i$.
To model high-order interactions among hops, we can stack $K$ hop interaction layers with residual connections as follows:
\begin{align}\label{eq:hopGNN}
    \mathbf{H}_i^k=\operatorname{GNN}(\mathcal{G}_i^\text{hop}, \mathbf{H}_i^{k-1}) + \mathbf{H}_i^{k-1}, \quad k=1\cdots K,
\end{align}
where $\mathbf{H}_i^k \in \mathbb{R}^{L \times d}$ is the $L$ hop features of node $v_i$ after the $k$-th hop interaction. In practice, we do not need to explicitly construct the hop feature graph $\mathcal{G}^\text{hop}$ for each node. 
Instead, without loss of generality, we implement the hop interaction $\operatorname{GNN}$ in Eq.~\eqref{eq:hopGNN} with a standard multi-head self-attention mechanism~\cite{GAT, vaswani2017attention} to capture the complex dependencies between hop features, \ie, pairwise interactions between two hop features in different semantic subspaces.
Specifically, the single head of self-attention on node $v_i$ is calculated as follows:
\begin{align}
&\mathbf{A}_i^\text{hop} = \operatorname{softmax}\left(\frac{\mathbf{H}_i\mathbf{W_Q}(\mathbf{H}_i\mathbf{W_K})^\top}{\sqrt{d}}\right),\\
&\operatorname{GNN}(\mathcal{G}^\text{hop}_i,\mathbf{H}_i) = \mathbf{A}^\text{hop}_i \mathbf{H}_i\mathbf{W_V}, 
\end{align} 
where we omit the superscript $k$ of layers for simplicity. Here, the $\mathbf{W_Q}$, $\mathbf{W_K}$, and $\mathbf{W_V}$ are learnable projection matrices in each interaction layer. The resulting $\mathbf{A}^\text{hop}_i$ can be interpreted as a weighted feature interaction graph that captures the interaction strength among multi-hop features.

Using the multi-head attention for hop interaction can be regarded as applying the GAT~\cite{GAT} as the hop interaction $\operatorname{GNN}$ over the hop feature graph $\mathcal{G}^\text{hop}_i$ in Eq~\eqref{eq:hopGNN}.
We also compared other GNN architectures~\cite{GCN,Hamilton2017InductiveRL} to model hop interactions in Section~\ref{sec:ablation} and found that they can also achieve comparable performance, which validates the generality and flexibility of our hop interaction framework.

\noindent \textbf{Hop Fusion and Prediction.}
After hop interaction, we apply a fusion and prediction function $\operatorname{g}$ for each node to generate the final output. We first apply a mean fusion to average all hop representations $\mathbf{H}^K \in \mathbb{R}^{N\times L \times d}$ into a single representation $\mathbf{Z} \in \mathbb{R}^{N \times d}$.
Here, we have tested other fusion mechanisms, but we did not observe noticeable improvements. Then, we utilize another simple linear layer with softmax activation for the final prediction $\mathbf{\hat{Y}} \in \mathbb{R}^{N \times c}$.
\begin{align}
\mathbf{Z} = \operatorname{fusion}(\mathbf{H}^K), \quad
\mathbf{\hat{Y}} = \operatorname{softmax}(\operatorname{linear}(\mathbf{Z})).
\end{align}
The training objective of HopGNN is the cross-entropy loss between the ground truth labels $\mathbf{Y}$ and the prediction $\mathbf{\hat{Y}}$:
\begin{equation}
    \mathcal{L}_\text{ce}= -\sum_{i \in \mathcal{V}_L} \sum_{j=1}^{c} \mathbf{Y}_{i j} \ln \mathbf{\hat{Y}}_{i j}.
\end{equation}
Although training HopGNN with only $\mathcal{L}_{\text {ce }}$ achieves competitive results, to show the generality and flexibility of our framework,  we will discuss how to combine HopGNN with self-supervised objectives~\cite{zbontar2021barlow,chen2020simple} in the following section.

\setlength{\tabcolsep}{4pt}
\begin{table*}[!t]
\scriptsize
    \centering  
    \caption{
Time and memory complexities. We consider time complexity for the feature propagation and transformation in the network and memory complexity for storing the activations of node embeddings (we ignore model weights since they are negligible compared to the activations). 
$L$ and $K$ are the number of hops and the non-linear transformation, respectively, and $d$ is the feature dimension (assumed to be fixed for all layers). $N$, $|E|$ and $s$ are the numbers of nodes, edges, and sampling neighborhoods, respectively. $b$ is the minibatch size.
	}
\renewcommand\arraystretch{1.2}
    \begin{adjustbox}{width=\textwidth}
\begin{tabular}{cc|c|c|ccc}
    \toprule
\multicolumn{2}{c|}{Category} & Method & Minibatch & Pre-Processing Time & Training Time & Training Memory \\ \hline
\multicolumn{2}{c|}{Standard} & GCN/GCNII/... & ${\times}$ & - & $O(LEd+LNd^2)$ & $O(LNd)$ \\ \hline
\multirow{4}{*}{Sampling} & Node & GraphSAGE & $\checkmark$ & $O(s^L N)$ & $O(s^L Nd^2)$ & $O(bs^Ld)$ \\
 & Layer & FastGCN/AS-GCN & $\checkmark$ & - & $O(sLNd^2)$ & $O(bsLd)$ \\
 & Graph & Cluster-GCN & $\checkmark$ & $O(E)$ & $O(LEd+LNd^2)$ & $O(bLd)$ \\
 & Graph & GraphSAINT & $\checkmark$ & $O(sN)$ & $O(LEd+LNd^2)$ & $O(bLd)$ \\ \hline
\multirow{3}{*}{Decoupling} 
& Pre & SGC/S2GC & $\checkmark$ & $O(LEd)$ & $O(Nd^2)$ & $O(bd)$ \\
& Pre & SIGN & $\checkmark$ & $O(LEd)$ & $O(KNd^2)$ & $O(bLd)$ \\
 & Post & APPNP/GPRGNN & ${\times}$ & - & $O(KNd^2+LEd)$ & $O(LNd)$ \\ \hline
\multicolumn{2}{c|}{Hop Interaction} & HopGNN & $\checkmark$ & $O(LEd)$ & $O(KNd^2+KNL^2d)$ & $O(bLd)$ \\ 
    \bottomrule
\end{tabular}
\end{adjustbox}
\label{tbl:complexity}
\end{table*}

\subsection{Self-Supervised Enhancement}

In this subsection, we show that it can easily incorporate the self-supervised learning (SSL) objective to further enhance the performance of HopGNN. Recall that the key idea of HopGNN is to conduct interactions among multi-hops to enhance the discrimination of nodes. The desired interaction would help nodes to capture the task-relevant features and drop task-irrelevant features. To encourage such property, recent feature-level SSL studies aim to maximize feature invariance between two views and minimize the redundancy between dimensions~\cite{zbontar2021barlow,zhang2021canonical}. Formally, in Barlow Twins~\cite{zbontar2021barlow}, given the cross-correlation matrix $\mathcal{C} \in \{-1,1\}^{d \times d}$ between two views, the SSL objective is:
\begin{align}\label{eq:ssl}
\mathcal{L}_{\text{ssl}} \triangleq \underbrace{\sum_i\left(1-\mathcal{C}_{i i}\right)^2}_{\text {invariance term }} + \quad \alpha \underbrace{\sum_i \sum_{j \neq i} \mathcal{C}_{i j}^2}_{\text {redundancy reduction term }},
\end{align}
where $\alpha$ is a scalar to control the decorrelation strength. Inspired by them, we further use multi-task learning to train the HopGNN, with the $\mathcal{L}_\text{ce}$ as the main task and $\mathcal{L}_\text{ssl}$ as the auxiliary task, and the overall training objective is: \begin{align}
    \mathcal{L}_{\text {final}}=\mathcal{L}_{\text {ce }}+\lambda \mathcal{L}_{\text {ssl }} .
\end{align}

\begin{figure}[t]
	\centering
	\includegraphics[width=1\linewidth]{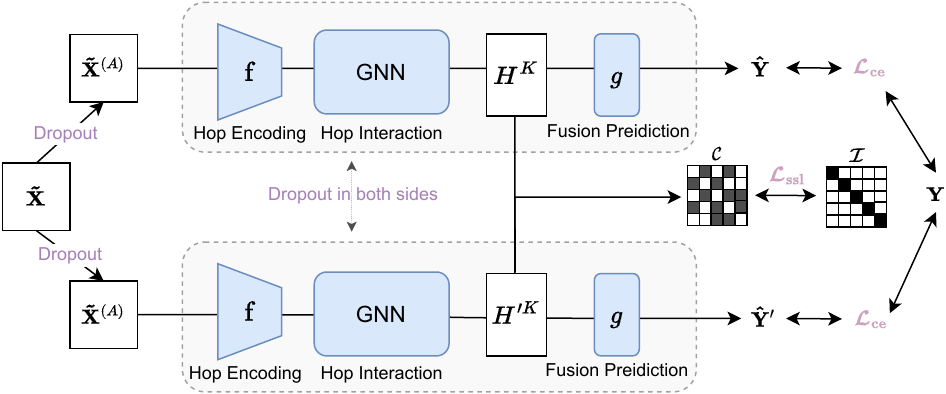}
	\caption{Multi-task learning with SSL for HopGNN. Illustration inspired by Barlow Twins~\protect\cite{zbontar2021barlow}.}\label{fig:ssl}
\end{figure}

In our case, as shown in Figure~\ref{fig:ssl}, the additional SSL objective is applied after the hop interaction to enhance the discrimination. To generate two views of hop interaction features, distinguished from the previous works to adopt sophisticated augmentation~\cite{chen2020simple, liu2022graph}, we simply forward HopGNN twice with different dropout units as augmentations to obtain $\mathbf{H}^K$ and $\mathbf{H}'^K$. 
Since the training of our model is independent of the graph structure, complex graph-based data augmentation operations are not needed.
Then, we not only calculate $\mathcal{L}_\text{ce}$ with fusion and prediction steps on both views but also flatten and normalize them to calculate the cross-correlation matrix $\mathcal{C}$ for $\mathcal{L}_\text{ssl}$.
Optimizing such the SSL objective can maximize the task-relevant mutual information and minimize the task-irrelevant information~\cite{zhang2021canonical}. It would help the learned hop interaction representation $\mathbf{H}^K$ to extract minimal and sufficient information about downstream tasks from multi-hop neighbors. 

Compared with other instance-level contrastive learning~\cite{chen2020simpleclr,khosla2020supervised}, the memory costs of the feature-level $\mathcal{L}_\text{ssl}$ do not increase with the graph size since it focuses on the feature dimension, which is scalable and efficient. We also provide the result of HopGNN with contrastive learning in the Appendix.

\subsection{Discussions}
\label{sec:method:discussion}
\noindent \textbf{Connection to decoupled GNNs.}
From the perspective of hop interaction, the decoupled GNNs learn a fixed linear combination of multi-hop features for all nodes, which is equivalent to applying fixed hop attention coefficients with diagonal parts and remaining off-diagonal as 0. However, such fixed coefficients ignoring the pairwise hop interaction would cause sub-optimal results since each hop's contribution for different nodes may be different. In contrast, our proposed HopGNN utilizes the self-attention mechanism to learn the representations of different-hop neighbors based on their semantic correlation for each node, which helps the model learn more informative node representations.

\noindent \textbf{Complexity analysis.}
Table \ref{tbl:complexity} provides a detailed asymptotic complexity comparison between HopGNN and other representative GNN methods. 1) The standard node interaction GNNs need full-batch training, and the time complexity contains the feature propagation part $O(LEd)$ over edges and feature transformation $O(LNd^2)$ over nodes. Moreover, for memory complexity, we need to store the activation of node embedding in each layer, which has $O(LNd)$. 
Note that we ignore the memory usage of model weights here since they are negligible compared to the activations~\cite{duan2022comprehensive}.
The time and memory complexity of full-batch node interaction GNNs are highly correlated to the size of the graphs and result in the scalability problem. 
2) Most sampling GNNs reduce the training time and memory cost via mini-batching with the corresponding sampling neighbor size $s$. 
3) For pre-computed decoupled GNNs, thanks to the pre-processing of feature propagation, the training complexity is the same as the traditional mini-batch training, \eg, MLPs with feature transformation, which is usually smaller than the sampling methods. However, the post-computed decoupled GNNs still require feature propagation during training, leading to a cost similar to that of full-batch GNNs.
4) The computational cost of HopGNN is similar to that of the pre-computed decoupled GNN, which is also scalable to large-scale graphs. Compared with SIGN, the HopGNN explicitly conducts non-linear interactions among $L$ hops, which is more expressive and requires $O(KNL^2d)$ additional time complexity of $K$ interactions.

\section{Experiments} 
\setlength{\tabcolsep}{3pt}
\begin{table*}[!t]
   \scriptsize
    \centering  
    \caption{Mean test accuracy $\pm$ stdev on 6 heterophily and 3 homophily real-world datasets over 10 public splits (48\%/32\%/20\% of nodes for training/validation/test). The best performance is highlighted. $\ddagger$ denotes the results obtained from previous works~\protect\cite{zhu2020beyond,yan2021two}}
    \label{tab:ssnc-results}
    \begin{adjustbox}{width=\textwidth}
    \begin{tabular}{l|cccccc|ccc|c} %
    \toprule
       & { Texas}& { Wisconsin}& { Actor} & { Squirrel} & { Chameleon} & { Cornell}& { Citeseer} &   { Pubmed} &   { Cora} & {{Avg}}\\
{$\mathcal{H}_\text{edge}$} & {0.11} & {0.21} & {0.22} & {0.22} & {0.23} & {0.3} & {0.74} & {0.8} & {0.81} &  -\\
{\#Nodes }  & {183} & {251} & {7,600} & {5,201} & {2,277} & {183} & {3,327} & {19,717} & {2,708} &  -\\
{\#Edges }  & {295} & {466} & {26,752} & {198,493} & {31,421} & {280} & {4,676} & {44,327} & {5,278} &  -\\
%{\#Features }  & {1703} & {1703} & {26,752} & {198,493} & {31,421} & {280} & {3,703} & {500} & {3,703} & - \\
{\#Classes }  & {5} & {5} & {5} & {5} & {5} & {5} & {7} & {3} & {6} & -\\
\midrule
	   {MLP$\ddagger$} & $81.89{\scriptstyle\pm4.78}$ & $85.29{\scriptstyle\pm3.61}$ & $35.76{\scriptstyle\pm0.98}$ & $29.68{\scriptstyle\pm1.81}$ & $46.36{\scriptstyle\pm2.52}$ & $81.08{\scriptstyle\pm6.37}$ &
	   $72.41{\scriptstyle\pm2.18}$ & $86.65{\scriptstyle\pm0.35}$ & $74.75{\scriptstyle\pm2.22}$ &  65.99\\
	   \midrule
	   GCN$\ddagger$ & $55.14 \scriptstyle\pm 5.16$ & $51.76 \scriptstyle\pm 3.06$ & $27.32 \scriptstyle\pm 1.10$ & $53.43 \scriptstyle\pm 2.01$ & $64.82 \scriptstyle\pm 2.24$ & $60.54 \scriptstyle\pm 5.30$ & $76.50 \scriptstyle\pm 1.36$ & $88.42 \scriptstyle\pm 0.50$ & $86.90 \scriptstyle\pm 1.04$ & 62.76  \\ 
	   GAT$\ddagger$ & $52.14 \scriptstyle\pm 5.16$ & $49.41 \scriptstyle\pm 4.09$ & $27.44 \scriptstyle\pm 0.89$ & $40.72 \scriptstyle\pm 1.55$ & $60.26 \scriptstyle\pm 2.50$ & $61.89 \scriptstyle\pm 5.05$ & $76.55 \scriptstyle\pm 1.23$ & $86.33 \scriptstyle\pm 0.48$ & $87.30 \scriptstyle\pm 1.10$ & 60.23  \\ 

	   {GraphSAGE$\ddagger$} & $82.43{\scriptstyle\pm6.14}$ & $81.18{\scriptstyle\pm5.56}$ & $34.23{\scriptstyle\pm0.99}$ & $41.61{\scriptstyle\pm0.74}$ & $58.73{\scriptstyle\pm1.68}$ & $75.95{\scriptstyle\pm5.01}$ & 
	   $76.04{\scriptstyle\pm1.30}$ & $88.45{\scriptstyle\pm0.50}$ & $86.90{\scriptstyle\pm1.04}$ & 69.50 \\
	   GCNII$\ddagger$ & $77.57 \scriptstyle\pm 3.83$ & $80.39 \scriptstyle\pm 3.40$ & $37.44 \scriptstyle\pm 1.30$ & $38.47 \scriptstyle\pm 1.58$ & $63.86 \scriptstyle\pm 3.04$ & $77.86 \scriptstyle\pm 3.79$ & \cellcolor{blue!15}$77.33 \scriptstyle\pm 1.48$ & $90.15 \scriptstyle\pm 0.43$ & \cellcolor{blue!15}$88.37 \scriptstyle\pm 1.25$ & 70.16  \\ 

	   \midrule
       {H2GCN-1$\ddagger$} & 
       $84.86{\scriptstyle\pm6.77}$ & 
       ${86.67}{\scriptstyle\pm4.69}$ & $35.86{\scriptstyle\pm1.03}$ & $36.42{\scriptstyle\pm1.89}$ & $57.11{\scriptstyle\pm1.58}$ & $82.16{\scriptstyle\pm4.80}$ &
       $77.07{\scriptstyle\pm1.64}$ & $89.40{\scriptstyle\pm0.34}$ & $86.92{\scriptstyle\pm1.37}$ & 70.72\\
      {H2GCN-2$\ddagger$} & 
      $82.16{\scriptstyle\pm5.28}$ & $85.88{\scriptstyle\pm4.22}$ & $35.62{\scriptstyle\pm1.30}$ & $37.90{\scriptstyle\pm2.02}$ & $59.39{\scriptstyle\pm1.98}$ & $82.16{\scriptstyle\pm6.00}$ &
      $76.88{\scriptstyle\pm1.77}$ & $89.59{\scriptstyle\pm0.33}$ & $87.81{\scriptstyle\pm1.35}$ & 70.87\\
ACM-GCN$\ddagger$ & \cellcolor{blue!15}$87.84 \scriptstyle\pm 4.40$ & \cellcolor{blue!15}$88.43 \scriptstyle\pm 3.22$ & $36.28 \scriptstyle\pm 1.09$ & $54.40 \scriptstyle\pm 1.88$ & $66.93 \scriptstyle\pm 1.85$ & $85.14 \scriptstyle\pm 6.07$ & $77.32 \scriptstyle\pm 1.70$ & $90.00 \scriptstyle\pm 0.52$ & $87.91 \scriptstyle\pm 0.95$ & 74.92  \\ 

    WRGAT$\ddagger$ & $83.62 \scriptstyle\pm 5.50$ & $86.98 \scriptstyle\pm 3.78$ & $36.53 \scriptstyle\pm 0.77$ & $48.85 \scriptstyle\pm 0.78$ & $65.24 \scriptstyle\pm 0.87$ & $81.62 \scriptstyle\pm 3.90$ & $76.81 \scriptstyle\pm 1.89$ & $88.52 \scriptstyle\pm 0.92$ & $87.95 \scriptstyle\pm 1.18$ & 72.90  \\ 
        GGCN$\ddagger$ & $84.86 \scriptstyle\pm 4.55$ & $86.86 \scriptstyle\pm 3.29$ & \cellcolor{blue!15}$37.54 \scriptstyle\pm 1.56$ & $55.17 \scriptstyle\pm 1.58$ & $71.14 \scriptstyle\pm 1.84$ & \cellcolor{blue!15}$85.68 \scriptstyle\pm 6.63$ & $77.14 \scriptstyle\pm 1.45$ & $89.15 \scriptstyle\pm 0.37$ & $87.95 \scriptstyle\pm 1.05$ & 75.05  \\ 
	   \midrule
S2GC & $68.65 \scriptstyle\pm 8.05$ & $71.57 \scriptstyle\pm 9.01$ & $34.17 \scriptstyle\pm 0.92$ & $41.63 \scriptstyle\pm 0.98$ & $58.55 \scriptstyle\pm 5.15$ & $75.25 \scriptstyle\pm 7.82$ & $76.08 \scriptstyle\pm 0.45$ & $88.31 \scriptstyle\pm 0.38$ & $87.73 \scriptstyle\pm 2.90$ & 66.88  \\ 
SIGN & $75.14 \scriptstyle\pm 7.94$ & $80.59 \scriptstyle\pm 3.75$ & $36.14 \scriptstyle\pm 1.01$ & $40.16 \scriptstyle\pm 2.12$ & $60.48 \scriptstyle\pm 2.10$ & $78.11 \scriptstyle\pm 4.67$ & $76.53 \scriptstyle\pm 1.76$ & $89.58 \scriptstyle\pm 0.45$ & $86.72 \scriptstyle\pm 1.37$ & 69.27  \\ 
       {APPNP} & $78.37{\scriptstyle\pm6.01}$ & $81.42{\scriptstyle\pm4.34}$ & $34.64{\scriptstyle\pm1.51}$ & $33.51{\scriptstyle\pm2.02}$ & $47.50{\scriptstyle\pm1.76}$ & $77.02{\scriptstyle\pm7.01}$ &
	   $77.06{\scriptstyle\pm1.73}$ & $87.94{\scriptstyle\pm0.56}$ & $87.71{\scriptstyle\pm1.34}$ &  67.24\\
	   {GPRGNN} & $82.12{\scriptstyle\pm7.72}$ & $81.16{\scriptstyle\pm3.17}$ & 
	   $33.29{\scriptstyle\pm1.39}$ & 
	   $43.29{\scriptstyle\pm1.66}$ & 
	   $61.82{\scriptstyle\pm2.39}$ &  
	   $81.08{\scriptstyle\pm6.59}$ &  
	   $75.56{\scriptstyle\pm1.62}$ & 
	   $86.85{\scriptstyle\pm0.46}$ & 
	   $86.98{\scriptstyle\pm1.33}$ & 70.15
	   \\ 
	   \midrule
HopGNN & $81.35 \scriptstyle\pm 4.31$ & $84.96 \scriptstyle\pm 4.11$ & $36.66 \scriptstyle\pm 1.39$ & $60.95 \scriptstyle\pm 1.65$ & $70.13 \scriptstyle\pm 1.39$ & $83.70 \scriptstyle\pm 6.52$ & $76.16 \scriptstyle\pm 1.53$ & $89.98 \scriptstyle\pm 0.39$ & $87.12 \scriptstyle\pm 1.35$ & 74.56  \\ 
HopGNN+ & $82.97 \scriptstyle\pm 5.12$ & $85.69 \scriptstyle\pm 5.43$ & $37.09 \scriptstyle\pm 0.97$ & \cellcolor{blue!15}$64.23 \scriptstyle\pm 1.33$ & \cellcolor{blue!15}$71.21 \scriptstyle\pm 1.45$ & $84.05 \scriptstyle\pm 4.48$ & $76.69 \scriptstyle\pm 1.56$ & \cellcolor{blue!15}$90.28 \scriptstyle\pm 0.42$ & $87.57 \scriptstyle\pm 1.33$ & \cellcolor{blue!15}75.53  \\ 
	   \bottomrule
    \end{tabular}
    \end{adjustbox}
    %\vspace{-0.5cm}
\end{table*}

\subsection{Setup}
\noindent \textbf{Datasets.}
We comprehensively evaluate the performance of HopGNN on 12 benchmark datasets. These datasets vary in domain, size and smoothness, including three standard homophily citation datasets~\cite{GCN}, six well-known heterophily datasets~\cite{pei2019geom}, two large-scale inductive datasets~\cite{zeng2019graphsaint} and a large-scale transductive OGB products dataset~\cite{hu2020open}.
For the homophily and heterophily benchmark datasets, we use the same 10 public fixed training/validation/test splits as provided in~\cite{pei2019geom,zhu2020beyond}. For large-scale datasets, we use their public splits in~\cite{zeng2019graphsaint, hu2020open}. The statistics of these datasets are summarized in Tables~\ref{tab:ssnc-results} and ~\ref{tab:large-scale}. Details on these datasets can be found in Appendix.

\noindent \textbf{Baselines.} 
We compare HopGNN with various baselines, including (1) MLP; (2) standard node-interaction GNN methods: GCN~\cite{GCN}, GAT~\cite{GAT}, GraphSAGE~\cite{Hamilton2017InductiveRL} and GCNII~\cite{chen2020simple}; (3) heterophilic GNNs with adaptive node interaction: H2GCN~\cite{zhu2020beyond},  WRGAT~\cite{suresh2021breaking}, ACM-GCN~\cite{luan2021heterophily}, GGCN~\cite{yan2021two} (4) sampling GNNs: FastGCN~\cite{chen2018fastgcn}, AS-GCN~\cite{huang2018adaptive}, ClusterGNN~\cite{chiang2019cluster}, GraphSAINT~\cite{zeng2019graphsaint}; and (5) decoupled GNNs: S2GC~\cite{zhu2020simple}, SIGN~\cite{frasca2020sign}, APPNP~\cite{klicpera2018predict}, GPRGNN~\cite{chien2020adaptive}. We report results from previous works with the same experimental setup if available.
If the results are not previously reported and codes are provided, we implement them based on the official codes and conduct a hyper-parameter search. 
We provide more baseline results in the Appendix due to space limits since.

\noindent \textbf{Implementation Details.}
Following the standard setting~\cite{hu2020open, bo2021beyond}, we set the hidden dimension of HopGNN as 128 for the nine small-scale datasets and 256 for the three large-scale datasets. 
Although tuning the hops and layers usually leads to better results, for simplicity, we fix the number of hops as 6 and the interaction layer as 2 of HopGNNs for all datasets in Table~\ref{tab:ssnc-results} and~\ref{tab:large-scale}.
We use Adam~\cite{kingma2014adam} for optimization and LayerNorm~\cite{Ba2016LayerN} of each layer and tune the other hyper-parameters. 
Details can be found in the Appendix.

\begin{table}[t]
{
\scriptsize
    \caption{
	 Comparison over a large-scale dataset. $\ddagger$ denotes the results obtained from previous works~\protect\cite{you2020design}.
	}
	\label{tab:large-scale}
	\renewcommand\arraystretch{1}
	\begin{adjustbox}{width=0.48\textwidth}
    \begin{tabular}{l|cccc}
    \toprule
     & {Flickr} & {Reddit} &  {Products}\\
{Type }  & {Inductive} & {Inductive} & {Transductive} \\
{\#Nodes }  & {89,250} & {232,965} &  {2,449,029}  \\
{\#Edges }  & {899,756} & {11,606,919} &  {61,859,140} \\
%\textbf{\#Features }  & {} & {1703} & {26,752} & {198,493} \\
{\#Classes }  & {7}& {41} & {47} \\

     \hline
    GCN$\ddagger$ & $49.2\pm 0.3$ & $93.3 \pm 0$ & $75.64\pm0.21$ \\
    SGC$\ddagger$ & $50.2\pm0.1$ & $94.9\pm0$ & $74.87\pm0.25$\\ 
    SIGN $\ddagger$ &  $51.4\pm0.1$ & $96.8\pm0$ & $77.60\pm0.13$\\
    S2GC & $50.48\pm 0.07$  & $94.04\pm0.03$ & $76.84\pm0.20$\\  
    \midrule
    GraphSAGE$\ddagger$ & $50.1\pm1.3$ & $95.3\pm0.1$ & $78.29\pm0.16$ \\
    FastGCN $\ddagger$& $50.4\pm0.1$ &$92.4\pm0.1$& $73.46\pm0.20$\\
    %Stochastic-GCN \\
    %LADIES & $50.51\pm0.13$ & $86.96\pm0.37$ & $75.31\pm0.56$\\ 
    AS-GCN$\ddagger$ & $50.4\pm0.2$ & $96.4\pm0.1$ & -\\
    ClusterGCN$\ddagger$ & $48.1\pm0.5$ & $95.4\pm0.1$ &$78.97\pm0.33$\\
    GraphSAINT$\ddagger$ &$51.1\pm0.1$ &$96.6\pm0.1$ &$79.08\pm0.24$\\
    \midrule
    HopGNN &$52.49 \pm 0.15$	&$96.92 \pm 0.05$ &$79.96 \pm 0.11$\\ 
    HopGNN+ &\cellcolor{blue!15}$52.68 \pm 0.16$ &\cellcolor{blue!15}$96.98 \pm 0.04$ &\cellcolor{blue!15}$80.08 \pm 0.08$\\ 
    \bottomrule
    \end{tabular}
        \end{adjustbox}
    }
\end{table}

\subsection{Overall Performance}

\noindent \textbf{Results on Homophily and Heterophily.}
From Table~\ref{tab:ssnc-results}, we make the following observations: 1) Standard node interactions are sometimes inferior to MLP, such as Actor and Cornell, indicating that simply stacking node interaction may fail in the heterophily datasets. However, the heterophilic GNNs achieve better performance in general, \eg, their average performance over nine datasets is all larger than 70. The reason is that such advanced node interaction can adaptively consider the influence of neighbors in each hop.
2) Compared with heterophilic GNNs, decoupled GNNs achieve sub-optimal results, \ie, their overall performance is less than 70, due to missing the non-linear interaction among nodes.
3) HopGNN achieves significantly better performance than decouple GNN and is comparable with the SOTA heterophilic GNNs. 
Such results show that even without complex node interactions, the non-linear interaction among hops can enhance node discrimination in both homophily and heterophily. This validates the effectiveness and generality of the hop interaction paradigm.
(4) Moreover, combining the SSL, the HopGNN+ consistently outperforms HopGNN and achieves the best overall performance, validating the effectiveness of the multi-task learning strategy and the compatibility of HopGNN.

\noindent \textbf{Results on Large-scale Datasets.}
We compare the HopGNN with scalable GNNs, including pre-processing decoupled GNNs and sampling-based GNNs in Table~\ref{tab:large-scale}, and find that: 1) The decoupled SIGN performs better in the inductive setting, and the subgraph-based GraphSAINT outperforms other baselines in the large-scale transductive product dataset. 2) HopGNN consistently outperforms all the baseline methods in these large-scale datasets. The results in the inductive setting of Flickr and Reddit substantiate the ability of HopGNN to generalize to unseen nodes. Moreover, the HopGNN+ combined feature-level SSL objective still works well in large-scale scenarios.

\subsection{Ablation Study}\label{sec:ablation}
In this part, we study the role of the hop-order embedding $\mathbf{E}_\text{order}$, hop interaction and hop fusion type to validate the effectiveness of each component and the generality of the whole framework. Due to space limits, we report the average test accuracy of these variants across three homophilic and six heterophilic datasets, as shown in Table~\ref{tab:ablation}.

\setlength{\tabcolsep}{5pt}
\begin{table}[]
\scriptsize
  \caption{
	 Ablation studies on order embedding, fusion and interaction types for heterophily and homophily graphs. 
	}
	\label{tab:ablation}
	\renewcommand\arraystretch{1.1}
	\begin{adjustbox}{width=0.48\textwidth}
\begin{tabular}{ll|ll|ll}
\toprule
\multicolumn{2}{l|}{}                               & \multicolumn{2}{l|}{Heterophily}  & \multicolumn{2}{l}{Homophily}   \\ \hline
\multicolumn{2}{l|}{HopGNN}                              & \multicolumn{2}{l|}{$69.63$} & \multicolumn{2}{l}{$84.42$} \\ \hline
\multicolumn{2}{l|} {w/o $\mathbf{E}_\text{order}$} & $60.93$         & {\color{red}  $\downarrow$}$8.70$  & $83.96$      & {\color{red}  $\downarrow$}$0.46$       \\ \hline
\multicolumn{1}{l|}{\multirow{2}{*}{Fusion}} & Attention   &    $69.81$           &   {\color{green}  $\uparrow$}$0.18$ &     $84.60$         &    {\color{green}  $\uparrow$}$0.18$       \\
\multicolumn{1}{l|}{}   & Max         &     $69.50$          &        {\color{red}  $\downarrow$}$0.13$    & $84.39$              &  {\color{red}  $\downarrow$}$0.03$           \\ \hline
\multicolumn{1}{l|}{\multirow{5}{*}{Interaction}}  & None        &    $65.59$           &  {\color{red}  $\downarrow$}$4.04$          &      $84.35$        &      {\color{red}  $\downarrow$}$0.07$     \\
\multicolumn{1}{l|}{}                        & MLP         &       $67.21$        &      {\color{red}  $\downarrow$}$2.42$     &      $84.41$        &     {\color{red}  $\downarrow$}$0.01$      \\
\multicolumn{1}{l|}{}                        & GCN         &        $68.18$       &      {\color{red}  $\downarrow$}$1.45$     &      $84.38$        &          {\color{red}  $\downarrow$}$0.04$  \\
\multicolumn{1}{l|}{}                        & SAGE        &        $69.11$       &  {\color{red}  $\downarrow$}$0.52$         &      $84.45$        &      {\color{green}  $\uparrow$}$0.03$      \\ \bottomrule
\end{tabular}
\end{adjustbox}
\end{table}

\noindent \textbf{Hop-order embedding.} 
Without hop-order embedding, the performance drops slightly in homophily (0.46) but dramatically in heterophily (8.7). This indicates that hop-order information is more crucial for order-insensitive hop interactions in challenging heterophilic scenarios. The reason may be that the contribution varies in different order hop of neighbors, which is also discussed in H2GNN~\cite{zhu2020beyond} that high-order neighbors are expected to be homophily dominant and useful for heterophilic datasets. 

\noindent \textbf{Hop fusion.} We also test the effects of max- and attention-based fusion mechanisms in HopGNN and observe their similar performance under both heterophily and homophily. Although the attention-based fusion achieves slightly higher results, we still choose the mean fusion as the default due to its simplicity and efficiency. 

\noindent \textbf{Hop interaction.} 
For the hop interaction layer, we test the variant of no interaction, interaction with GCN and SAGE, and interaction with MLP. We make the following observations: 1) They perform closely under homophily since the nodes' neighbors share similar information, resulting in limited gain from interactions.
2) Removing the interaction incurs a performance reduction of 4.04 in heterophily datasets, and the interaction based on MLP also results in a 2.3 performance drop. These results are consistent with the limited performance of decoupled GNNs using linear combination or concatenation among hops. Such performance degeneration further validates the importance of hop-level interaction since it can investigate discriminative clues of nodes for the classification in heterophily. 3) For the interaction of GNN variants, both standard GCN and SAGE achieve competitive results, demonstrating that our framework is stable with different GNN-based hop interaction mechanisms. We choose the multi-head attention mechanism as the default due to its generalizability.

\begin{figure}[!t]
	\includegraphics[width=1\linewidth]{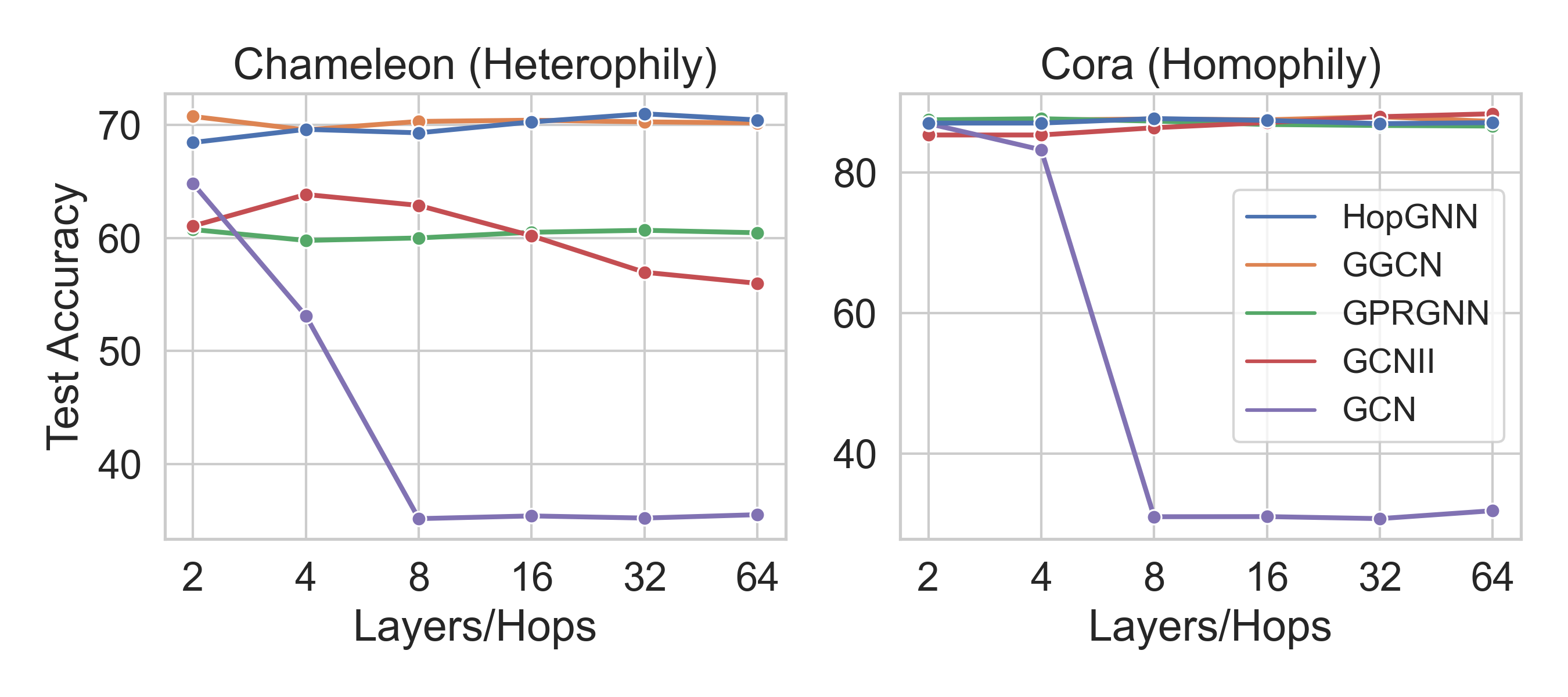}
\caption{Classification accuracy with different layers/hops.}
\label{fig:num_hops}
\end{figure}

\subsection{In-depth Analysis}
\noindent \textbf{Over-smoothing.}
To test the robustness of the models to over-smoothing, we compared the HopGNN with the classical GCN and different kinds of SOTA, including GRPGNN from decoupled GNNs, GCNII from deep GNNs, and GCNN from heterophilic GNNs. From Figure~\ref{fig:num_hops}, we have the following observations: 1) The performance of the GCN drops rapidly as the number of layers increases. However, all the other models do not suffer over-smoothing in the homophilic datasets. 2) Both GCNII and GPRGNN significantly underperform the GGCN and HopGNN under all layers in Chameleon. This means that although decoupling and residual connection can solve the over-smoothing to some extent, they are insufficient for heterophily. 
3) The GGCN conducts adaptive node interactions in each layer by carefully considering each hop neighbor's information, which is expressive but limits its scalability. Notability, without node interaction, the HopGNN still achieves robustness under different layers in both homophily and heterophily, validating the superiority of the hop interaction paradigm.

\begin{figure}[!t]
	\centering
	\includegraphics[width=1\linewidth]{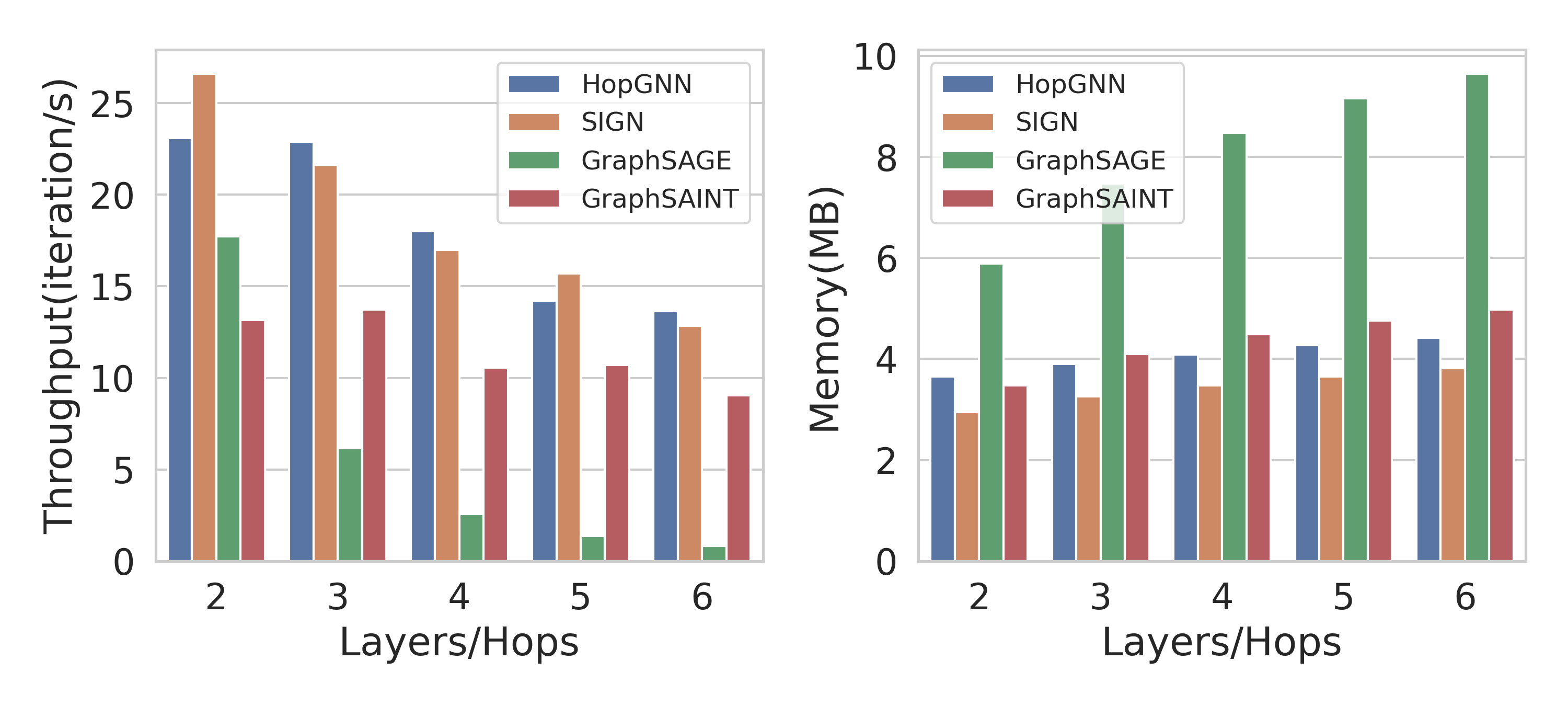}
	\caption{The comparison of {\it throughput} (left) and {\it memory usage} (right) in the Products dataset, both y-axis are in the log scale.}
	\label{fig:efficiency}
\end{figure}

\noindent \textbf{Efficiency.}
In Figure~\ref{fig:efficiency}, we use the evaluation methodology from~\cite{duan2022comprehensive} to fairly compare the throughputs and actual memory usage of various representative methods. Specifically, we compare the HopGNN with SAGE variants under the same settings across layers during the training procedure on the largest Products dataset. We observe that: 1) The basic GraphSAGE is significantly slower and costs a huge amount of memory due to the neighbor explosion. Sub-graph sampling makes GraphSAINT significantly reduce time and memory costs. 2) The decoupled SIGN achieves faster and smaller memory cost than GraphSAINT, but with sub-optimal results as shown in Table~\ref{tab:large-scale}. 3) HopGNN costs slightly more memory than SIGN due to the additional hop interaction phase, but HopGNN achieves the best performance and is faster than GraphSAINT, which indicates a better trade-off between performance and scalability. 
The implementation details and training efficiency comparison can be found in Appendix.

\section{Conclusion}
We have presented a novel hop interaction paradigm, a practical solution to address the scalability and over-smoothing problem of GNNs simultaneously. 
It shifts the interaction target of GNNs from nodes to hops inside each node.
Specifically, we design a simple yet effective HopGNN framework. It first pre-computes non-parameter aggregation of multi-hop features to reduce the computational cost during training and inference. 
Then, it conducts non-linear interactions among the multi-hop features of each node via GNNs to enhance their discriminative abilities. 
We also develop a multi-task learning strategy with the self-supervised objective to enhance the performance of the downstream task.
Experiments on 12 benchmark datasets show that HopGNN achieves state-of-the-art performance while maintaining high scalability and efficiency. 
One future work is to investigate advanced interaction mechanisms among hops to enhance the performance of HopGNNs. It is also interesting to apply the HopGNN to other downstream tasks, such as graph classification and link prediction.

\clearpage

%%%%%%%%% REFERENCES
{\small
\bibliographystyle{ieee_fullname}
\bibliography{egbib}
}
\appendix
\section{Appendix}
\subsection{Details of Datasets}
We provide the details of datasets in the following:

\begin{itemize}
    \item {Homophily Datasets}
    \begin{itemize}
        \item \textit{Citeseer, Pubmed, Cora}~\cite{GCN}: For the benchmark citation datasets, nodes correspond to papers, edges correspond to citation links, the sparse bag-of-words are the features, and each node's label represents the paper's topic. 
    \end{itemize}
    \item {Heterophily Datasets}
    \begin{itemize}
        \item \textit{Texas, Wisconsin, Cornell}~\cite{pei2019geom}: Nodes and edges represent the web pages and hyperlinks captured from the computer science departments of these universities in the WebKB dataset. Nodes' features are bag-of-word representations of contents on these web pages. Each node is labeled into five categories: student, project, course, staff, and faculty.
        \item \textit{Squirrel, Chameleon}~\cite{pei2019geom}: Chameleon and Squirrel are web pages extracted from different topics in Wikipedia. Nodes and edges denote the web pages and hyperlinks among them, respectively, and informative nouns in the web pages are employed to construct the node features in the bag-of-word form. Webpages are labeled in terms of the average monthly traffic level.
        \item \textit{Actor}~\cite{pei2019geom}: The actor network contains the co-occurrences of actors in films, which describes the complex relationships among films, directors, actors and writers. In this network, nodes and edges represent actors and their co-occurrences in films, respectively. The actor’s Wikipedia page is used to extract features and node labels.
    \end{itemize}
        \item {Large-scale Datasets}
    \begin{itemize}
        \item \textit{Flickr, Reddit}~\cite{Hamilton2017InductiveRL,zeng2019graphsaint}: Flickr and Reddit are inductive datasets for multiclass categorization. The goal of Reddit is to predict online post communities based on user comments. The task of Flickr is to classify the categories of images based on their description and common characteristics of online photographs.
        \item \textit{Products}~\cite{hu2020open}: The Products dataset is a large-scale Amazon product co-purchasing network in a transductive setting. Nodes represent products sold in Amazon, edges indicate the products purchased together, and features are 100-dimensional product descriptions after Principal Component Analysis. Labels are the category of products.
    \end{itemize}
\end{itemize}

\subsection{Implementation Details}

\noindent \textbf{Hyper-parameters.} For HopGNN, we search the hyper-parameters, including learning rate from [0.01, 0.001, 0.005], weight decay from [0, 5e-4, 5e-5, 5e-6], dropout rate from [0.2, 0.4, 0.5, 0.6], $\alpha$ from [0.01, 0.1, 0.5, 0.8] and $\lambda$ from [1e-4, 5e-4] via validation sets of each dataset. 
For other baselines, we search the layers/hops from [2, 8, 16, 32] and fix the hidden dimensions as 128. The search space of other hyper-parameters, such as learning rate, weight decay, and dropout, is the same as that of HopGNN.

\noindent \textbf{Hardware and Environment.} We run our experiments on a single machine with 
%Dual Intel Xeon CPUs (E5-2698 v4 @ 2.2Ghz),
Intel Xeon CPUs (Gold 5120 @ 2.20GHz),
one NVIDIA Tesla V100 GPU (32GB of memory) and 512GB DDR4 memory. We use PyTorch 1.11.0 with CUDA 10.2 to train the model on GPUs.

\noindent \textbf{Throughput and Memory.} For fairness, we set the hidden dimension to 256 and control the batchsize as 3000 across different models on the Products dataset. We report the hardware throughput and activation usage-based on~\cite{duan2022comprehensive}. The throughput measures how many times each model can complete training steps within a second. We measure the activation memory using $\textit{torch.cuda.memory\_allocated}$. 

\subsection*{A.3. More Experiment Results}

\setlength{\tabcolsep}{3pt}
\begin{table*}[t]
   \scriptsize
    \centering  
    \caption{Mean test accuracy $\pm$ stdev. The best performance is highlighted. $\ddagger$ denotes the results obtained from previous works~\protect\cite{zhu2020beyond,yan2021two}.}
    \label{tab:ssnc-results2}
    \begin{adjustbox}{width=\textwidth}
    \begin{tabular}{l|cccccc|ccc|c} %
    \toprule
       & { Texas}& { Wisconsin}& { Actor} & { Squirrel} & { Chameleon} & { Cornell}& { Citeseer} &   { Pubmed} &   { Cora} & {{Avg}}\\
\midrule
	   {GCN+JK$\ddagger$} & $66.49{\scriptstyle\pm6.64}$ & $74.31{\scriptstyle\pm6.43}$ & $34.18{\scriptstyle\pm0.85}$ & $40.45{\scriptstyle\pm1.61}$ & $63.42{\scriptstyle\pm2.00}$ & $64.59{\scriptstyle\pm8.68}$ & $74.51{\scriptstyle\pm1.75}$ & $88.41{\scriptstyle\pm0.45}$ & $86.79{\scriptstyle\pm0.92}$ & 65.79\\
	   {GCN-Cheby$\ddagger$} & $77.30 \scriptstyle\pm 4.07$ & $79.41 \scriptstyle\pm 4.46$ & $34.11 \scriptstyle\pm 1.09$ & $43.86 \scriptstyle\pm 1.64$ & $55.24 \scriptstyle\pm 2.76$ & $74.32 \scriptstyle\pm 7.46$ & $75.82 \scriptstyle\pm 1.53$ & $88.72 \scriptstyle\pm 0.55$ & $86.76 \scriptstyle\pm 0.95$ & 68.39  \\ 

	   {MixHop$\ddagger$} & $77.84{\scriptstyle\pm7.73}$ & $75.88{\scriptstyle\pm4.90}$ & $32.22{\scriptstyle\pm2.34}$ & $43.80{\scriptstyle\pm1.48}$ & $60.50{\scriptstyle\pm2.53}$ & $73.51{\scriptstyle\pm6.34}$ & $76.26{\scriptstyle\pm1.33}$ & $85.31{\scriptstyle\pm0.61}$ & \cellcolor{blue!15}${87.61}{\scriptstyle\pm0.85}$ & 68.21\\
	       {GEOM $\ddagger$} & $67.57$ & $64.12$ & $31.63$ & $38.14$ & $60.90$ & $60.81$  & \cellcolor{blue!15}${77.99}$ &  ${90.05}$ & $85.27$ & 64.05 \\
	   %{SGC} \\ 
	   {FAGCN} & $78.11 {\scriptstyle\pm 5.01}$ & $81.56{\scriptstyle\pm4.64}$ & $35.41{\scriptstyle\pm1.18}$ & $42.43{\scriptstyle\pm2.11}$ & $56.31{\scriptstyle\pm3.22}$ & $76.12{\scriptstyle\pm7.65}$ & $74.86{\scriptstyle\pm2.42}$ &  $85.74{\scriptstyle\pm0.36}$ & $83.21{\scriptstyle\pm 2.04}$ & 68.18  \\
	   {DAGNN} & $70.27{\scriptstyle\pm4.93}$ & $71.76{\scriptstyle\pm5.25}$ &$35.51{\scriptstyle\pm1.10}$ &$30.29{\scriptstyle\pm2.23}$ & $45.92{\scriptstyle\pm2.30}$ &	$73.51{\scriptstyle\pm7.18}$  &$76.44{\scriptstyle\pm1.97}$ 	&$89.37{\scriptstyle\pm0.52}$ & $86.82{\scriptstyle\pm1.67}$ & 64.43\\
	   \midrule
HopGNN & $81.35 \scriptstyle\pm 4.31$ & $84.96 \scriptstyle\pm 4.11$ & $36.66 \scriptstyle\pm 1.39$ & $60.95 \scriptstyle\pm 1.65$ & $70.13 \scriptstyle\pm 1.39$ & $83.70 \scriptstyle\pm 6.52$ & $76.16 \scriptstyle\pm 1.53$ & $89.98 \scriptstyle\pm 0.39$ & $87.12 \scriptstyle\pm 1.35$ & 74.56  \\ 
HopGNN+ & \cellcolor{blue!15}$82.97 \scriptstyle\pm 5.12$ & \cellcolor{blue!15}$85.69 \scriptstyle\pm 5.43$ & \cellcolor{blue!15}$37.09 \scriptstyle\pm 0.97$ & \cellcolor{blue!15}$64.23 \scriptstyle\pm 1.33$ & \cellcolor{blue!15}$71.21 \scriptstyle\pm 1.45$ & 
$84.05 \scriptstyle\pm 4.48$ & $76.69 \scriptstyle\pm 1.56$ & \cellcolor{blue!15}$90.28 \scriptstyle\pm 0.42$ & $87.57 \scriptstyle\pm 1.33$ & \cellcolor{blue!15}75.53  \\ 
HopGNN+SCL & $81.65 \scriptstyle\pm 7.47$ & $84.37 \scriptstyle\pm 4.91$ & $36.72 \scriptstyle\pm 1.05$ & $61.42 \scriptstyle\pm 1.98$ & $70.45 \scriptstyle\pm 1.03$ & \cellcolor{blue!15}$84.32 \scriptstyle\pm 6.71$ & $76.59 \scriptstyle\pm 1.51$ & $90.01 \scriptstyle\pm 0.29$ & $87.28 \scriptstyle\pm 1.71$ & 74.76  \\ 

	   \bottomrule
    \end{tabular}
    \end{adjustbox}
    %\vspace{-0.5cm}
\end{table*}

\noindent \textbf{More Baselines.} We provide more baseline results in Table~\ref{tab:ssnc-results2}, including 1) the classical advanced node interaction with high-order neighbor information: GCN+JK~\cite{xu2018representation}, GCN-ChebyNet~\cite{defferrard2016convolutional}, and MixHop~\cite{abu2019mixhop}. 2) the heterophilic GNNs: Geom-GCN~\cite{pei2019geom} and FAGCN~\cite{bo2021beyond} and 3) decoupled GNN: DAGNN~\cite{liu2020towards}. However, their average performance on nine datasets is less than 70, indicating that they have been shown to be outperformed by the SOTA methods.

\noindent \textbf{Supervised Contrastive Learning Objective.} As discussed in~\cite{zhang2021unleashing}, the supervised contrastive loss~\cite{khosla2020supervised} (SCL) may help to regularize the feature space and make it more discriminative, \ie, minimizing the SCL is equivalent to minimizing the class-conditional entropy $\mathcal{H}(\mathbf{H}|\mathbf{Y})$ and maximizing the feature entropy $\mathcal{H}(\mathbf{H})$. Therefore, it can also be used to enhance the discriminative ability of hop interaction, and we provide the results of HopGNN with SCL in Table~\ref{tab:ssnc-results2}. Formaly, the SCL objective can be defined as:

\begin{align}
\mathcal{L}_\text{final} &=\mathcal{L}_\text{CE}+\lambda \mathcal{L}_\text{SCL}, \\
\mathcal{L}_\text{SCL} &=\sum_{i \in {I}_{\mathcal{L}}} \frac{-1}{|P(i)|} \sum_{p \in P(i)} \log \frac{\exp \left(\mathbf{h}_i \cdot \mathbf{h}_p / \tau\right)}{\sum_{a \in A(i)} \exp \left(\mathbf{h}_i \cdot \mathbf{h}_a / \tau\right)},
\end{align}
where $i \in {I}_{\mathcal{L}} \equiv\{1 \ldots 2 b\}$ is the index of an arbitrary augmented sample within a training batch (size=b), and $A(i) \equiv {I}_{\mathcal{L}} \backslash \{i\}$. $P(i) \equiv\left\{p \in A(i): {\mathbf{y}}_p={\mathbf{y}}_i\right\}$ is the set of indices of all positives that share same label y in a multiview batch distinct from $i$, and $|P(i)|$ is its cardinality. 

As shown in the last row of Table~\ref{tab:ssnc-results2}, the result demonstrates that the SCL can also improve the performance of HopGNN and is comparable with HopGNN+, which validates the compatibility of the HopGNN framework to different SSL objectives. 
However, the complexity of SCL is $O(N^2)$ in a batch due to the instance discrimination task. Therefore, the SCL is not as scalable as the feature-level SSL used in HopGNN+.

\begin{figure}[!t]
	\includegraphics[width=1\linewidth]{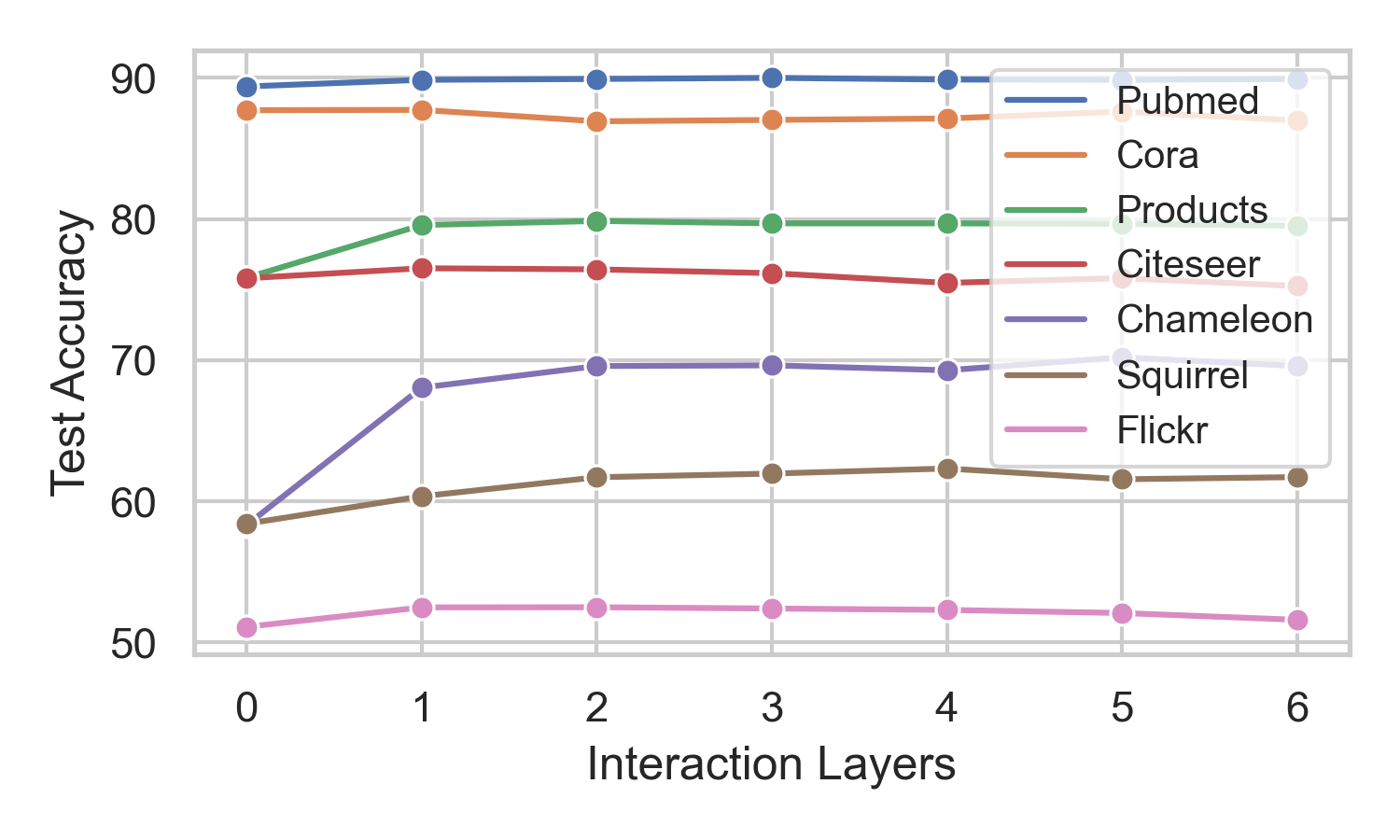}
\caption{Classification accuracy with different interaction layers.}
\label{fig:num_inters}
\end{figure}

\noindent \textbf{The Effects of Interaction Layers.}
We test the effects of different interaction layers in HopGNN under various datasets, including Cora, Citeseer, and Pubmed as homophily examples; Chameleon and Squirrel as heterophily examples; Flickr and Products as large-scale examples. In Figure~\ref{fig:num_inters}, we observe that HopGNN achieves competitive results using two interaction layers in most cases. Thus, we adopt the default layers of hop interaction as two.
Compared with homophily datasets, increasing the layers of hop interaction would remarkably boost the performance of heterophily datasets. The reason may be that such high-order hop interactions may capture the co-occurrence pattern of multi-hop neighbors, which contain the discriminative clues for heterophily.

\begin{figure}[!t]
  %\fbox{\rule{0pt}{0.5in} \rule{0.9\linewidth}{0pt}}
  \includegraphics[width=1\linewidth]{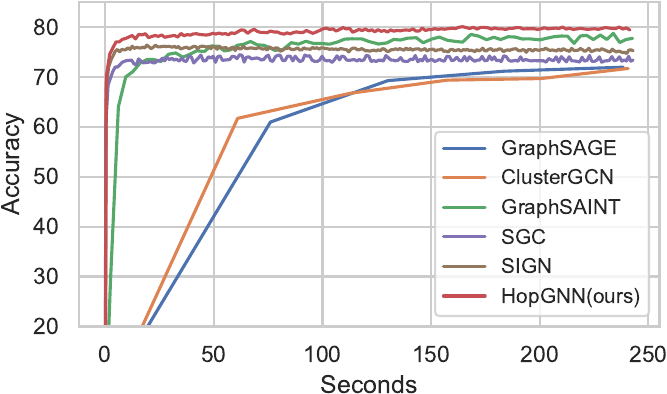}
   \caption{{Empirical efficiency study on the largest Product dataset: \textit{x}-axis shows the total training time (previous 250 seconds), \textit{y}-axis is the accuracy on the test set.}}
   \label{fig:empirical}
   \vspace{-0.6cm}
\end{figure}

\noindent \textbf{Training Efficiency.}
In Figure~\ref{fig:empirical}, we have added experiments with representative baselines. HopGNN converges most quickly and performs best empirically, which also validates that our hop interaction framework can achieve a better trade-off between effectiveness and efficiency.

\begin{figure*}[!ht]
	\centering
	\includegraphics[width=\linewidth]{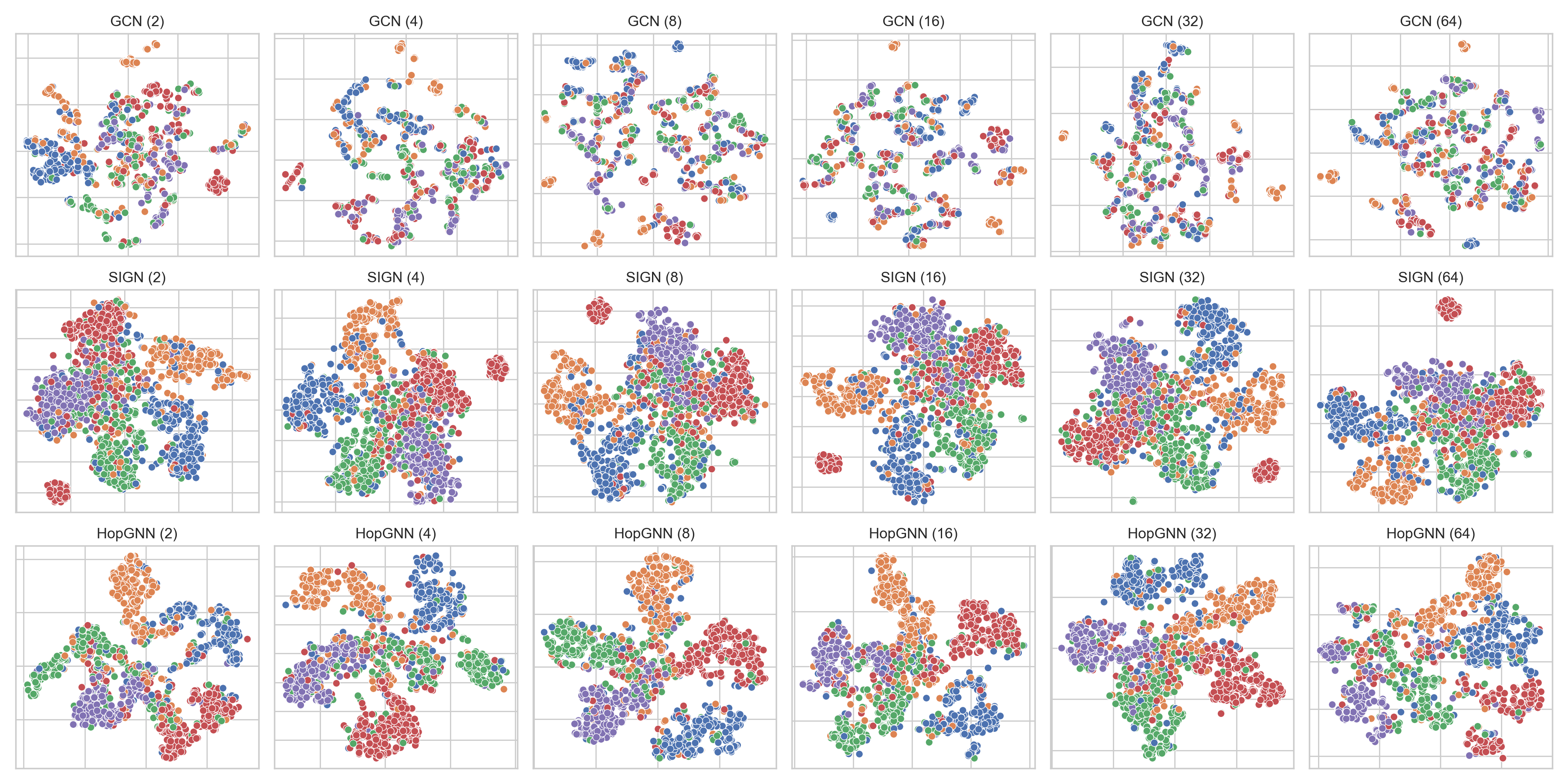}
	\caption{ t-SNE visualization of node representations derived by different numbers of layers/hops of models on \textbf{Chameleon}. Colors represent node classes, and the number in the bracket indicates the layers/hops. Note that these clusters correspond to the five classes, indicating that HopGNN has better discriminative power under different layers in heterophily.}
	\label{fig:tsne}
\end{figure*}

\begin{figure*}[!h]
	\centering
	\includegraphics[width=\linewidth]{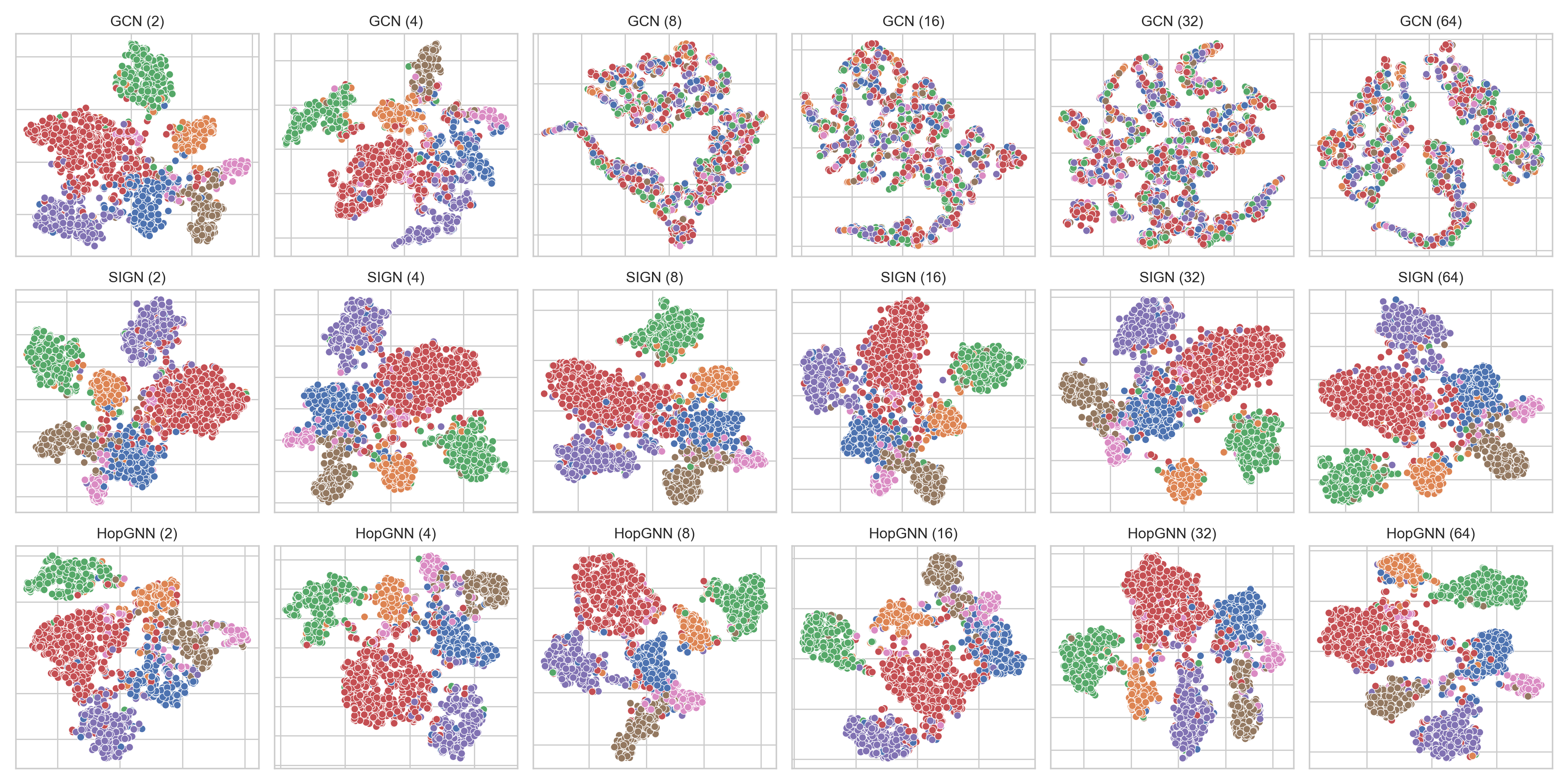}
	\caption{ t-SNE visualization of node representations derived by different numbers of layers/hops of models on \textbf{Cora}. Compared with GCN, the representation of both SIGN and HopGNN are discriminative when increasing layers in a homophily scenario. However, due to the simplicity of homophily datasets, the SIGN and HopGNN achieve similar results.}
	\label{fig:tsne_cora}
\end{figure*}

\noindent \textbf{Visualization of Representation.} 
To qualitatively investigate the effectiveness of the learned feature representation of nodes, we provide a visualization of the t-SNE for the last layer features of GCN (node interaction), SIGN (decoupled), and HopGNN (hop interaction) on the Chameleon (Figure~\ref{fig:tsne}) and Cora (Figure~\ref{fig:tsne_cora}). Compared with GCN (first row), the SIGN (second row) and HopGNN (last row) are both robust when increasing layers under the Chameleon and Cora. Moreover, the representation of HopGNN exhibits more discernible clustering in Chameleon. Note that these clusters correspond to the five classes of the dataset, verifying the better discriminative power of HopGNN under different layers in Heterophily.

\end{document}